\documentclass[runningheads]{llncs}

\usepackage{makeidx}  
\usepackage{url}
\usepackage{color}
\usepackage{longtable}
\usepackage{multirow}
\usepackage{graphicx}
\usepackage{amssymb, amsmath, bm}
\usepackage{mathtools}
\usepackage{mathrsfs}
\usepackage{booktabs}
\usepackage{cite}
\usepackage{dsfont}
\usepackage[colorlinks,linkcolor=blue]{hyperref}
\usepackage{makecell}
\usepackage[noend]{algpseudocode}
\usepackage{algorithmicx,algorithm}
\usepackage{graphicx}

\usepackage[misc]{ifsym} 
\usepackage{bbding}
\usepackage[colorlinks,linkcolor=blue]{hyperref}
\usepackage{makecell}
\usepackage[noend]{algpseudocode}
\usepackage{algorithmicx,algorithm}
\usepackage{graphicx}
\usepackage[misc]{ifsym} 
\usepackage{bbding}
\newcommand{\repeatthanks}{\textsuperscript{\thefootnote}}
\begin{document}
\title{Fine-grained Correlation Loss for Regression}
%
%
\author{Chaoyu Chen\inst{1,2,3}\thanks{Chaoyu Chen and Xin Yang contribute equally to this work.} \and Xin Yang\inst{1,2,3}\repeatthanks \and Ruobing Huang\inst{1,2,3} \and Xindi Hu\inst{4} \and Yankai Huang\inst{5} \and Xiduo Lu\inst{1,2,3} \and Xinrui Zhou\inst{1,2,3} \and Mingyuan Luo\inst{1,2,3} \and Yinyu Ye\inst{1,2,3} \and Xue Shuang\inst{1,2,3} \and Juzheng Miao\inst{6} \and Yi Xiong\inst{5}\and Dong Ni\inst{1,2,3}\textsuperscript{(\Letter)}} 

\institute{
\textsuperscript{$1$}National-Regional Key Technology Engineering Laboratory for Medical Ultrasound, School of Biomedical Engineering, Health Science Center, Shenzhen University, China\\
\email{nidong@szu.edu.cn} \\
\textsuperscript{$2$}Medical Ultrasound Image Computing (MUSIC) Lab, Shenzhen University, China\\
\textsuperscript{$3$}Marshall Laboratory of Biomedical Engineering, Shenzhen University, China\\
\textsuperscript{$4$}Shenzhen RayShape Medical Technology Co., Ltd, China\\
\textsuperscript{$5$}Department of Ultrasound, Luohu People’s Hosptial, Shenzhen, China\\
\textsuperscript{$6$}School of Biological Science and Medical Engineering, Southeast University, China}
\authorrunning{C.Chen et al.}
\maketitle
\begin{abstract}
Regression learning is classic and fundamental for medical image analysis. It provides the continuous mapping for many critical applications, like the attribute estimation, object detection, segmentation and non-rigid registration. However, previous studies mainly took the case-wise criteria, like the mean square errors, as the optimization objectives. They ignored the very important population-wise \textit{correlation} criterion, which is exactly the final evaluation metric in many tasks. In this work, we propose to revisit the classic regression tasks with novel investigations on directly optimizing the fine-grained correlation losses. We mainly explore two complementary correlation indexes as learnable losses: Pearson linear correlation (PLC) and Spearman rank correlation (SRC). The contributions of this paper are two folds. First, for the PLC on global level, we propose a strategy to make it robust against the outliers and regularize the key distribution factors. These efforts significantly stabilize the learning and magnify the efficacy of PLC. Second, for the SRC on local level, we propose a coarse-to-fine scheme to ease the learning of the exact ranking order among samples. Specifically, we convert the learning for the ranking of samples into the learning of similarity relationships among samples. We extensively validate our method on two typical ultrasound image regression tasks, including the image quality assessment and bio-metric measurement. Experiments prove that, with the fine-grained guidance in directly optimizing the correlation, the regression performances are significantly improved. Our proposed correlation losses are general and can be extended to more important applications.
\end{abstract}

\section{Introduction}
\begin{figure*}
    \centering
    \includegraphics[width=0.9\linewidth]{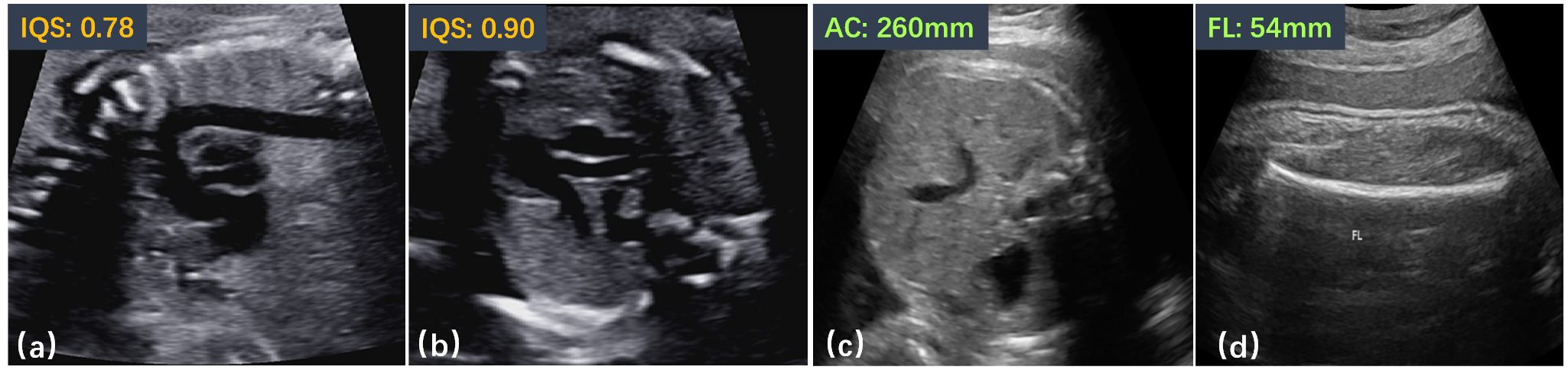}
    \caption{The regression tasks considered in this paper. (a)-(b): IQA on fetal heart US images with IQS stands for image quality score. (c)-(d): AC and FL denote abdominal circumference and femur length of fetus in US images, respectively.}
    \label{fig:tasks}
\end{figure*}

Regression is a statistical method that attempts to determine the continuous mapping relationship between one dependent 
variable and another one. Regression has been explored as a fundamental solution for versatile medical image analysis, 
e.g. image quality assessment (IQA)\cite{gao2020combined}, landmark localization \cite{payer2016regressing,noothout2020deep}, 
object detection\cite{ding2017accurate}, segmentation \cite{he2019fully}, bio-metric measurement (BMM)\cite{zhang2020direct} 
and registration \cite{cao2018deformable}. Fig.\ref{fig:tasks} shows the tasks we accomplished with regression in this paper, 
including IQA and bio-metric estimation in fetal ultrasound (US) images. \par

Lots of approaches have been devised for the regression. Dong et al. \cite{han2014robust} proposed a regression forest to predict the landmark coordinates in brain MR. Wang et al. \cite{wang2014regression} formulated the spine segmentation as a boundary regression. In deep learning era, Cao et al. \cite{cao2018deformable} designed deep networks to regress the deformation field. Korhonen et al. \cite{you2021transformer} built a deep transformer for IQA regression. In recent years, researchers explored to introduce extra information for case-wise supervision and improve the regression. Christian et al. \cite{payer2016regressing} proposed to regress the Gaussian heatmap instead of coordinates for landmark detection. In \cite{noothout2020deep}, authors collected both landmark and classification cues to refine the location regression. \par

To the best of our knowledge, these medical regression studies mainly focus on learning the mapping among input and output for individual samples, but ignore the learning of the structured relationships over the dataset and among the samples. \textit{Correlation} criterion at population level just fits the very capacities to describe the relationships, including linear relationship and ranking order. Recent work in computer vision community explored to directly optimize the correlation based objectives. V-MEON-SF \cite{liu2018end} and Norm-In-Norm \cite{li2020norm} approximated the Pearson linear correlation (PLC) based loss function on global level for regression. Although effective, their solution is sensitive to the outliers. Further, Engilberge et al. \cite{engilberge2019sodeep} use deep networks to directly regress the absolute ranking order with brute force. The solution is heavy and still overlooks the relative ranking among samples. \par
In this work, we propose to directly optimize the correlation in a fine-grained way to improve medical image regression. We mainly explore two complementary correlation indexes with novel formulations for maximization: PLC to regularize the strength of linear relationship and Spearman rank correlation (SRC) to further emphasize relative ranking order. Our contributions are two folds. First, for the PLC on global level, we propose a new design by customizing different optimization objectives for normal and outliers. Further, we propose to directly regularize the key distribution factors for regression. These composite efforts significantly stabilize the learning and strengthen the efficacy of PLC. Second, for the SRC on local level, we propose a lightweight, coarse-to-fine scheme to ease the learning of the relative ranking order among samples. Specifically, we transform the learning into the constraints on similarity relationships among samples. We extensively validate our method on two typical ultrasound image regression tasks, including the IQA and BMM. Experiments prove that, with the novel formulations in directly optimizing the correlation, our method can significantly improve the regression performances. Our proposed method is general and can be considered in more applications.\par
\begin{figure}
    \centering
    \includegraphics[width=0.9\textwidth]{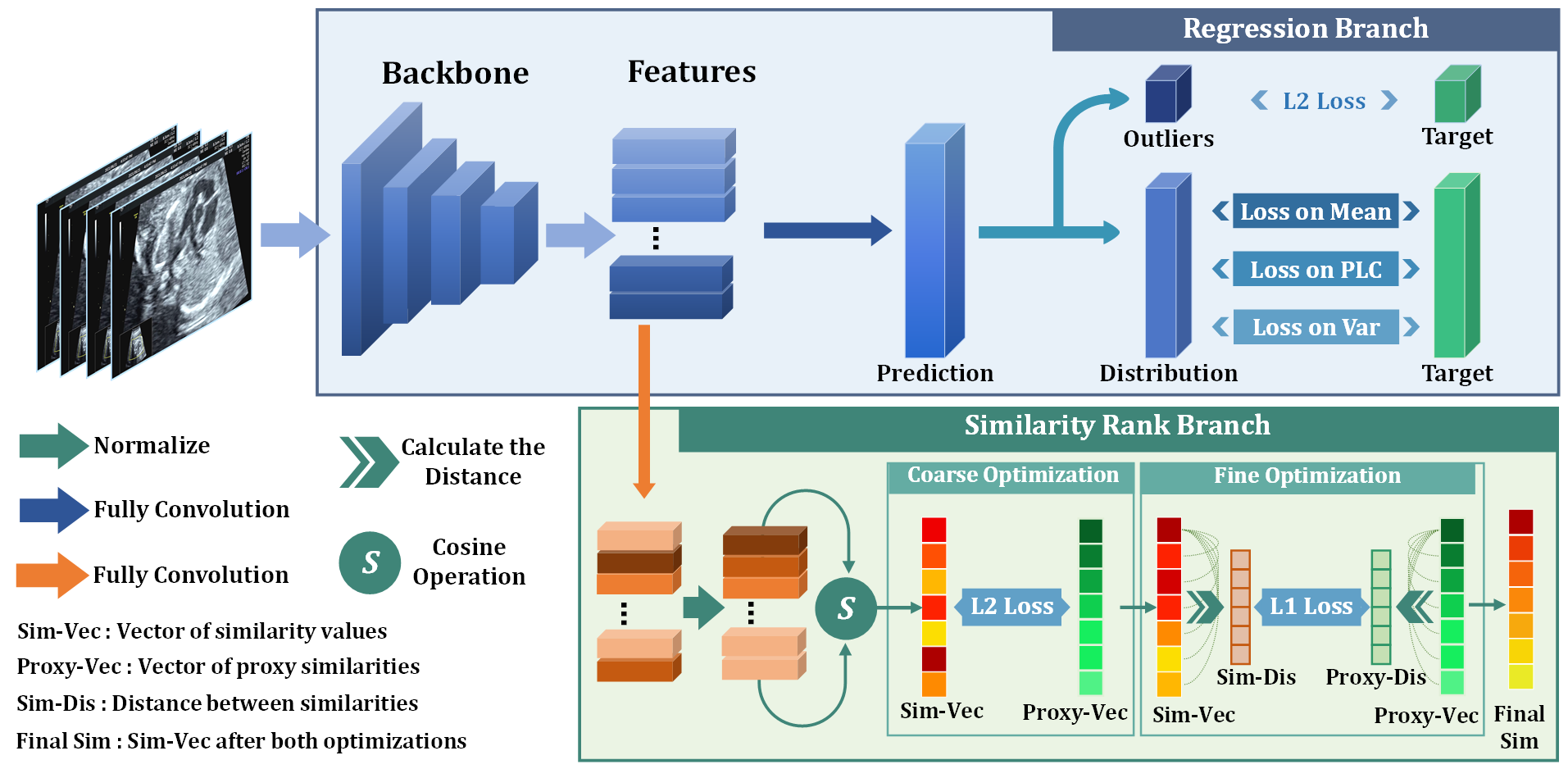}
    \caption{Schematic view of our proposed method.} 
    \label{framework}
\end{figure}

\section{Methodology}
Fig.~\ref{framework} provides an overview of our framework in the training phase. For a batch of image samples, the backbone firstly extracts feature tensors and then forwards them to the PLC based regression branch. The branch automatically identifies the outliers from normal samples and splits the batch into two parts. The normal part then receives the supervision from PLC loss. The feature tensors also flow into the similarity rank branch to constrain the ranking order. In testing, the regression branch would provide the final prediction. \par

\subsection{Effective PLC Loss to Optimize Linear Relationship}
PLC is used to measure the strength of linear relationship between two variables. We aim to take PLC as the objective to maximize and hence reduce the distribution discrepancy at global level. The basic definition of PLC between $X$ and $Y$ follows the Pearson index as follows: 
\begin{equation}
\small
    PLC(X,Y) = \frac{\sum_{i=1}^{n}({X_i-\mu _{X}})({Y_i-\mu _{Y}})}{\sqrt{\sum_{i=1}^{n}({X_i-\mu _{X}})^2} \sqrt{\sum_{i=1}^{n}({Y_i-\mu _{Y}})^2}},
    \label{pearson_func}
\end{equation}
where \begin{math}\mu _{X}\end{math} or \begin{math}\mu _{Y}\end{math} denotes the mean value of the variable. This definition is inherently very sensitive to the outliers in the prediction. The outliers mean the samples with predictions differ severely from the ground truth. Therefore, we propose to automatically identify and handle the outliers with customized loss for a stable training. For the rest clean samples, we move further to narrow the constraints on the distribution by minimizing the distance between means and variances, respectively. Finally, as Fig. \ref{framework} illustrates, the detailed definition of our reformed PLC loss is:
\begin{equation}
    \small
    \begin{aligned}
    Loss_{PLC} = \left\{
        \begin{matrix}
        &\frac{1}{m} \sum_{i=1}^{m} (\hat{y}_{i} - y_{i} )^{2}, \text{ if $y_{i}$ is outlier,}
        \\\\
        &1 - PLC^{2}(\hat{Y},Y) + (\mu _{\hat{Y}} - \mu _{Y})^2 + (\sigma  _{\hat{Y}} - \sigma_{Y})^2, \space \mathrm{otherwise}
        \end{matrix}\right.
    \end{aligned}
    \label{pearson_loss}
\end{equation}

where \begin{math}\hat{y}\in \hat{Y}\end{math} and \begin{math}y\in Y\end{math} represent the ground truth and the regression result, respectively. $\sigma$ is the variance of variable. We take the top 10$\%$ samples with the largest difference between $\hat{y}$ and $y$ in each iteration as outliers. \par
\begin{figure}
    \centering
    \includegraphics[width=0.95\textwidth]{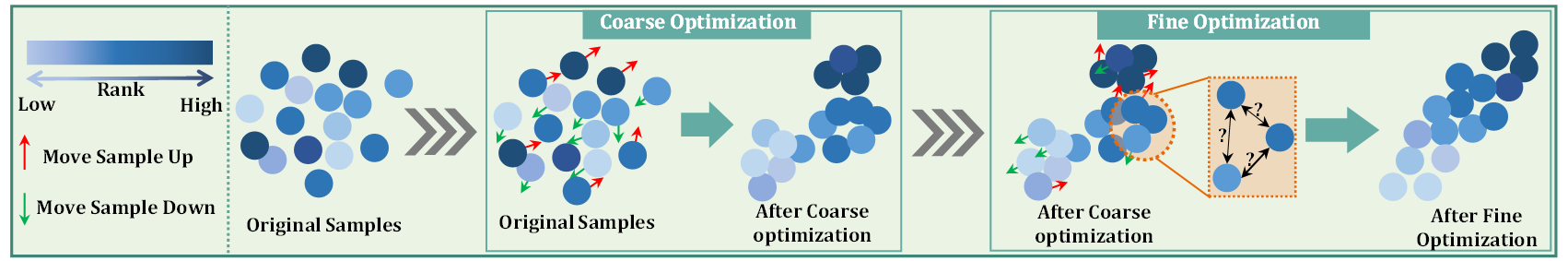}
    \caption{The coarse-to-fine rank learning scheme.} 
    \label{fig:Sample_rank}
\end{figure}
\subsection{Coarse-to-Fine SRC to Regularize Rank}
A perfect regression should simultaneously satisfy the criteria from individual, global and peer levels. Previous methods mainly focus on optimizing the losses at individual level, like the mean square error, but ignore the correlation relationship. Our reformed PLC loss partially addresses the problem at global level, but still misses an important relationship at peer level, i.e., the ranking order. Our proposed SRC loss aims to solve this problem. The original SRC is defined to measure the strength of association between two ranked variables as $SRC(\mathrm{X} ,\mathrm{Y}) = 1 - (6||\mathbf{rk}(\mathrm{X})-\mathbf{rk}(\mathrm{Y})||^2_2)/(n(n^2-1)) $, where \begin{math}\mathrm{X}\end{math} and \begin{math}\mathrm{Y}\end{math} denote two variables, \begin{math}\mathbf{rk}(\cdot)\end{math} is the sorting operator,  \begin{math}n\end{math} is the sample number of the variable. Low $SRC$ means $X$ and $Y$ matches each other on not only the value but also the strict rank. However, the discrete ranking involved loss is difficult to minimize, like the SRC which is non-differentiable in its original formulation. Moreover, how to describe the absolute and relative ranking for learning is challenging. \par

In this paper, we propose to transform the learning of the ranking order into the problem of regularizing the similarities among samples. This transformation provides not only a differentiable approximation for SRC, but also a new way for rank representation. Specifically, the sample similarity is represented by the cosine distance $\mathbf{S}(x_i,x_j)$ between feature vectors (Fig. \ref{framework}). Notably, as shown in Fig. \ref{fig:Sample_rank}, to further ease the rank learning, we decompose the rank among samples into global and local rank and propose a coarse-to-fine learning scheme (Fig. \ref{framework}).

\subsubsection{Coarse Level for Global Rank.}
In the coarse level, as Fig. \ref{fig:Sample_rank} shows, we aim to roughly regularize the global rank by adding constraints on sample similarity. Our hypothesis is that the samples with similar rank ground truth should have similar feature representations to some extent, while the images with dissimilar ranks should be pushed away from each other. We propose $\mathbf{P}(x_i,x_j)=\left[\frac{\mathbf{R}(x_i)}{\mathbf{R}(x_j)} \right ]$ as a proxy similarity reference between sample $i$ and $j$, where $x_i$ is the feature vector, $\mathbf{R}(\cdot)$ denotes the vector's regression target. $\left [ \cdot \right ]$ is the reciprocal operation when the value is greater than 1.0. Based on our hypothesis and $\mathbf{P}(x_i,x_j)$, we can conduct the rough regularization on features to encourage the clustering and scattering of samples guided by the $\mathbf{R}(\cdot)$ (Fig. \ref{fig:Sample_rank}). The sample similarity $\mathbf{S}(x_i,x_j)$ should satisfy the following condition in this coarse level:
\begin{equation}
r(x_i,x_j):\mathbf{P}(x_i,x_j)<
|\mathbf{S}(x_i,x_j)|<\mathbf{P}(x_i,x_j) + \alpha 
\label{mapping_relaship}
\end{equation}
The parameter $\alpha $ is an interval and controls the margin. We set $\alpha $ to 0.25 in this study. Then, the first part of our SRC loss is defined as Eq. \ref{Eq:spearman_loss_part1} for rough penalty:
\begin{equation}
 L_C= (|\mathbf{S}(x_i, x_j)| - \mathbf{P}(x_i,x_j))^2, \text{ if $|\mathbf{S}(x_i, x_j)|$ not subject to $r(x_i, x_j)$.} 
 \label{Eq:spearman_loss_part1}
\end{equation}

\subsubsection{Fine Level for Local Rank.}
$L_C$ in coarse level set rough cluster and rank for samples. However, the exact rank in local range among samples is not reviewed. Therefore, when Eq. \ref{mapping_relaship} is satisfied, we further propose to regularize the local rank for any tuple of sample features $(x_i,x_j,x_k)$ in a fine level, as the triangle relationship shown in Fig. \ref{fig:Sample_rank}. Our intuitive motivation is that, if the target rank of the tuple $(x_i,x_j,x_k)$ presents the relationship $\mathbf{R}(x_i) > \mathbf{R}(x_j) > \mathbf{R}(x_k)$, then the similarity metrics should accordingly satisfy the symmetric distance relationship of $\mathbf{S}(x_i,x_j) > \mathbf{S}(x_i,x_k)$ and $\mathbf{S}(x_j,x_k) > \mathbf{S}(x_i,x_k)$. Therefore, we can realize the sample ranking by constraining the similarities. \par

Inspired by the margin-based ranking loss \cite{schroff2015facenet}, the difference between the similarities should be larger than a margin, in order to better distinguish the relationship between similar samples. Instead of using a fixed margin in \cite{schroff2015facenet}, we propose to build an adaptive margin to better fit the diverse similarity relationships in regression task. Hence, we introduce the difference between the proxy similarity reference $\mathbf{P}(x_i,x_j)$ as the dynamic margin. We therefore build the following losses as the second part of our SRC to regularize the local rank:
\begin{equation}
    L_{ascent} = max\left \{ 0, \left (\mathbf{P}(x_i,x_j) - \mathbf{P}(x_i,x_k) \right)
    - \left (|\mathbf{S}(x_i, x_j)| - |\mathbf{S}(x_i, x_k)|   \right ) \right \} 
    \label{spearman_loss_part2_pos}
    \end{equation}
\begin{equation}
    L_{descent} = max\left \{ 0, \left (\mathbf{P}(x_j,x_k) - \mathbf{P}(x_i,x_k)\right ) - \left (|\mathbf{S}(x_j, x_k)| - |\mathbf{S}(x_i, x_k)|   \right )\right \}  
    \label{spearman_loss_part2_neg}
    \end{equation}
The tuple $(x_i,x_j,x_k)$ in Eq. \ref{spearman_loss_part2_pos} and Eq. \ref{spearman_loss_part2_neg} is in its sorted version according to the ground truth rank. We considered both $L_{ascent}$ and $L_{descent}$ as a bi-directional design for symmetric constraints on similarity relationships. Ablation studies in Tab. \ref{abl_table} validates the importance of this design. Finally, our SRC loss covering global and local rank, with coarse-to-fine learning scheme, is defined as:
\begin{equation}
    Loss_{SRC} = \left\{\begin{matrix} 
        {\frac{1}{m}}\sum_{(i,j,k)}^{m} (L_{ascent} + L_{descent}),\space \mathrm{otherwise}
         \\\\
        {\frac{1}{n}}\sum_{(i,j)}^{n}L_{C}, \text{ if $|\mathbf{S}(x_i, x_j)|$ not subject to $r(x_i, x_j)$.} 
        \end{matrix}\right. 
        \label{spearman_loss}
\end{equation}

\section{Experimental Results}
\subsection{Materials and Implementation Details}
We evaluate the regression performance of the proposed framework on two typical ultrasound image-based regression tasks. The first one is IQA, consisting of 2000 images of right ventricular outflow tract (RVOT), 2000 images of pulmonary bifurcation (PB) and 4475 images of aortic arch (AA). The second task is BMM, including 5026 images of fetal head circumference (HC), 4551 images of fetal abdominal circumference (AC) and 6454 images of femur length measurement (FL). The pixel resolution of AC, HC and FL is $0.22\times 0.22mm^2$, $0.18\times0.18mm^2$ and $0.17\times0.17mm^2$, respectively. Image quality score of IQA and measurement length of BMM were manually annotated by experts using the Pair annotation software package\cite{liang2022sketch}. We randomly split the dataset into a training ($60\%$), a validation ($15\%$) and a test set ($25\%$). All experimental results are based on three-fold cross-validation and presented in the form of mean(std). \par
We implement our method in Pytorch and train the system by Adam optimizer, using a standard server with an NVIDIA 2080Ti GPU. For optimization, we run 160 epochs with a stage-wise learning rate: initial value of 3e-4 and decaying to $25\%$ of previous for every 70 epochs of training for all experiments. The batch size is set to 160 on all datasets. In addition, we employ a warm-up strategy to train the network with a basic loss function (i.e. Mean Square Error). Then, we introduce the correlation-based loss functions when the training meets both of the following conditions: 1) training for more than 30 epochs; 2) the PLC and SRC on the validation did not rise for 5 consecutive epochs. Data augmentation includes random rotation and random scaling. The input size of the image is $320\times320$ and the fixed size is generated by padding after scaling. \par
\subsection{Quantitative and Qualitative Analysis}
The performance is measured via three criteria for IQA task: PLC $\uparrow$, SRC $\uparrow$ and Kendall correlation (KLC $\uparrow$), and another two criteria are used for BMM task: absolute error (AE $\downarrow$) and relative error (RE $\downarrow$). We use Resnet18 as the network for all methods. Ablation study was conducted by comparing the methods including Resnet18, Resnet18 with PLC loss (Res-PLC), Resnet18 with SRC loss (Res-SRC) and Resnet18 with PLC and SRC loss (Res-PLC-SRC). We also compared with NIN~\cite{li2020norm} and SoDeep~\cite{engilberge2019sodeep} under the same warm-up strategy. \par
\begin{table}
    \centering
    \scriptsize
    \caption{Quantitative evaluation of mean(std) results for two regression tasks.}\label{main_result_table}

    \begin{tabular}{c | c |c |c| c| c| c| c}
    \toprule[1pt]
    \multicolumn{2}{c|}{Methods}  & Resnet18  & Res-PLC & NIN\cite{li2020norm}  & Res-SRC & SoDeep\cite{engilberge2019sodeep}   &Res-PLC-SRC  \\ 
    \hline
    \multirow{3}*{ROVT} &  PLC     & 0.597(0.01) & 0.616(0.02) & 0.616(0.03) & 0.602(0.01) &  0.574(0.01) & $\textcolor{blue}{0.659(0.01)}$  \\ 
                        &  SRC    & 0.551(0.02) & 0.651(0.03) & 0.628(0.01) & 0.583(0.02)   &  0.670(0.02) & $\textcolor{blue}{0.685(0.01)}$  \\ 
                        &  KLC     & 0.398(0.01) & 0.477(0.02) & 0.456(0.01) & 0.422(0.01)  &  0.490(0.01) & $\textcolor{blue}{0.506(0.00)}$  \\ 
    \hline
            
    \multirow{3}*{PB} &  PLC        & 0.622(0.04) & $\textcolor{blue}{0.662(0.03)}$ & 0.570(0.03) &0.627(0.07) & 0.565(0.08) & 0.649(0.06) \\ 
                        &  SRC     & 0.583(0.05) & 0.615(0.04) & 0.596(0.07) & 0.627(0.07) & 0.585(0.07) & $\textcolor{blue}{0.648(0.05)}$  \\ 
                         &  KLC     & 0.420(0.04) & 0.444(0.03) & 0.425(0.06) & 0.445(0.06)  & 0.418(0.06) & $\textcolor{blue}{0.469(0.04)}$ \\ 
    \hline
    \multirow{3}*{AA} &  PLC       & 0.771(0.01) & 0.808(0.01) & 0.804(0.01) & 0.801(0.01) & 0.786(0.01) & $\textcolor{blue}{0.820(0.01)}$ \\ 
                    &  SRC        & 0.796(0.03) & 0.821(0.02) & 0.814(0.02) & 0.813(0.01) & 0.810(0.03) & $\textcolor{blue}{0.832(0.02)}$ \\ 
                    &  KLC         & 0.610(0.02) & 0.632(0.02) & 0.630(0.02) & 0.628(0.01) & 0.626(0.03) & $\textcolor{blue}{0.642(0.01)}$ \\ 
    \hline
    \hline       
    \multirow{5}*{AC} &  PLC     & 0.977(0.03) & 0.980(0.02) &  0.962(0.03) & 0.978(0.02) &  0.966(0.03)& $\textcolor{blue}{0.984(0.02)}$ \\ 
                        &  SRC     & 0.955(0.03) & 0.963(0.02)&  0.961(0.03)&0.962(0.02) &  0.961(0.03)& $\textcolor{blue}{0.968(0.02)}$  \\ 
                        &  KLC     & 0.855(0.07) & 0.869(0.06) &  0.860(0.07)&0.863(0.06) &  0.867(0.04)& $\textcolor{blue}{0.878(0.06)}$  \\ 
                        &  AE($mm$)   & 5.74(6.38) & 6.01(6.28)&  68.36(6.67)&6.53(4.90) &  39.18(5.08)& $\textcolor{blue}{5.47(5.51)}$ \\ 
                        &  RE($\%$)    & 4.21(5.59) & 4.43(5.54) &  42.94(3.42)&5.23(5.89) &  32.74(18.85)& $\textcolor{blue}{3.86(4.73)}$ \\ 
    
    \hline
    \multirow{5}*{HC} &  PLC      & 0.995(0.00)& 0.997(0.00)&  0.994(0.00)&0.996(0.00) &  0.988(0.01)& $\textcolor{blue}{0.998(0.00)}$ \\ 
                        &  SRC     & 0.993(0.01)& 0.995(0.00)&  0.993(0.00)&0.995(0.00)&  0.992(0.01)& $\textcolor{blue}{0.996(0.00)}$  \\ 
                        &  KLC     & 0.943(0.03)& 0.953(0.02) &  0.933(0.01)&0.947(0.02) &  0.945(0.03)& $\textcolor{blue}{0.954(0.02)}$  \\ 
                        &  AE($mm$)  & 2.07(1.04)& 1.55(1.09)&  32.44(2.24)& 1.73(1.48) &  25.96(3.06)& $\textcolor{blue}{1.52(1.26)}$\\ 
                        &  RE($\%$)    & 2.64(1.72) & 1.96(1.51) &  27.41(4.13)&2.40(2.69) &  39.61(12.54)& $\textcolor{blue}{1.85 (1.42)}$ \\ 
    \hline

    \multirow{5}*{FL} &  PLC      & 0.975(0.02)& 0.979(0.02)&  0.978(0.01)&0.977(0.01) &  0.972(0.01)& $\textcolor{blue}{0.982(0.01)}$ \\ 
                        &  SRC     & 0.963(0.01)& 0.973(0.01)&  0.972(0.00)&0.970(0.01)&  0.966(0.02)& $\textcolor{blue}{0.978(0.01)}$  \\ 
                        &  KLC     & 0.850(0.03)& 0.877(0.03) &  0.873(0.04)&0.868(0.03)&  0.883(0.02)& $\textcolor{blue}{0.890(0.03)}$  \\ 
                        &  AE($mm$)   & 1.73(1.04)& 1.27(1.20)&  14.55(1.49)&1.32(1.32) &  5.63(1.04)& $\textcolor{blue}{1.17(1.22)}$ \\ 
                        &  RE($\%$)    & 11.28(11.63)& 8.09(11.82) &  38.59(2.56)&9.40(15.98) &  33.56(19.39)& $\textcolor{blue}{7.02(11.32)}$\\ 
                    
    \toprule[1pt]
    \end{tabular}
    \end{table}

Table~\ref{main_result_table} compares the performance of the different methods on two regression tasks.  Comparing Res-PLC and NIN\cite{li2020norm} can prove that paying attention to outliers and explicitly constraining the mean and variance of the distribution can significantly improve model performance. Our proposed Res-SRC outperforms SoDeep\cite{engilberge2019sodeep} in overall comparison demonstrates that the Coarse-to-Fine strategy is more suitable for learning rank. Moreover, SoDeep and NIN perform poor in AE and RE is because they only focus on the rank correlation among predictions and labels, while ignore the value differences. This results in high correlation values, but poor AE and RE. According to Table~\ref{main_result_table}, the superior performance of Res-PLC and Res-SRC compared to Resnet18 shows the efficacy of our proposed PLC loss and SRC loss separately. Furthermore, the Res-PLC-SRC achieved the best result on both regression tasks among all these methods further demonstrates the effectiveness and generality of our approach. \par

\begin{figure}
    \scriptsize
    \centering
    \includegraphics[width=1.0\linewidth]{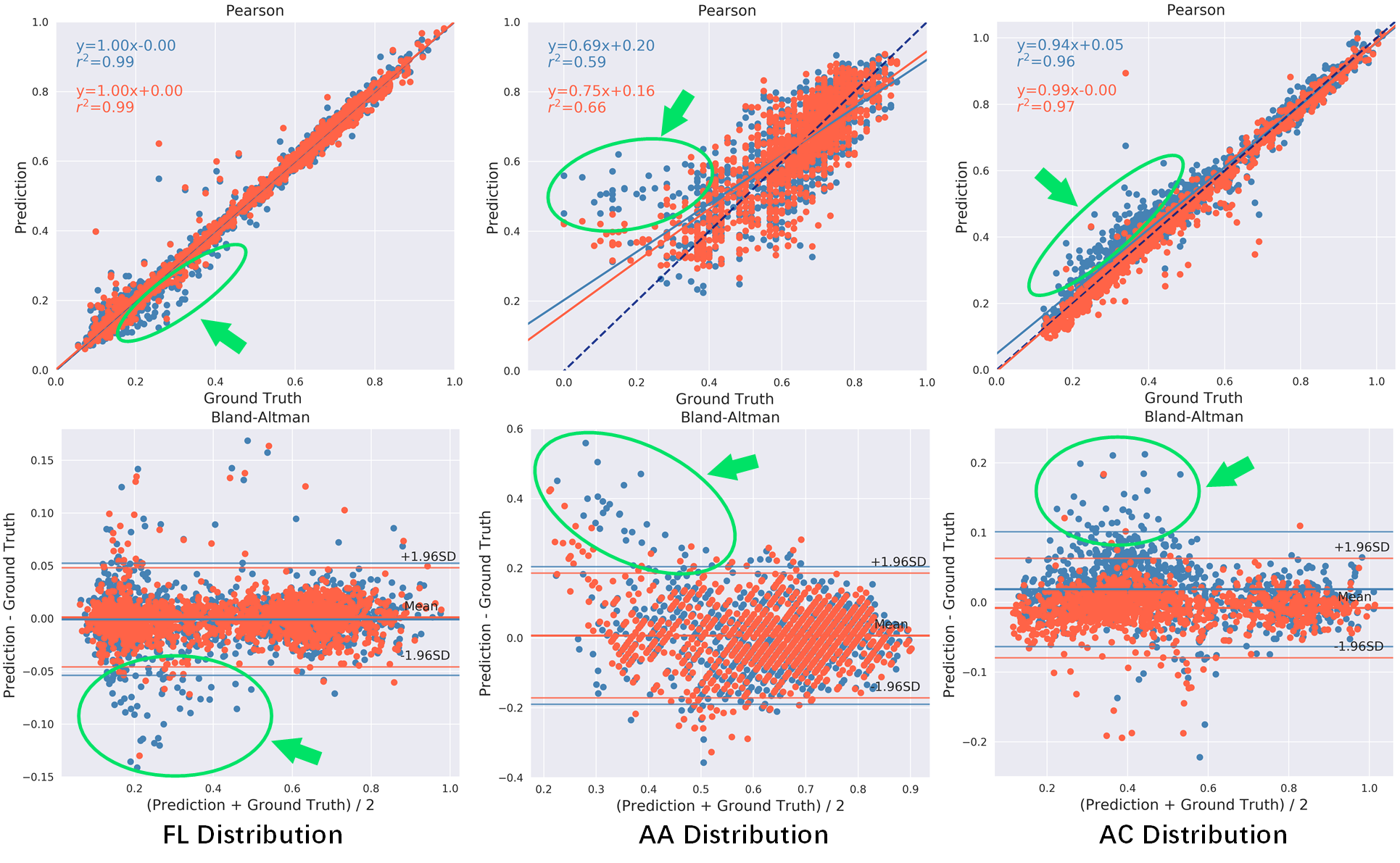}
    \caption{Visualization of correlation and consistency.
    Blue dots are the Resnet18's result and red dots are our approach's result. Green circles indicate outliers.} \label{distribute_fig}
    \end{figure}

\begin{table}
    \centering
    \caption{Ablation Results of SRC loss.}\label{abl_table}

    \begin{tabular}{c|c|c|c|c|c}
    \toprule[1pt]
    \multicolumn{2}{c|}{Methods}  & Resnet18  & Res-$\mathrm{SRC_{ascent}}$ & Res-$\mathrm{SRC_{descent}}$ &Res-SRC  \\ 

    \hline
            
    \multirow{3}*{ROTV} &  PLC     & 0.597(0.01) & 0.583(0.02)  & 0.562(0.01)  & $\textcolor{blue}{0.602(0.01)}$ \\ 
                        &  SRC     & 0.551(0.02) & 0.569(0.01) &0.559(0.06) & $\textcolor{blue}{0.583(0.02)}$ \\ 
                         &  KLC     & 0.398(0.01) & 0.403(0.01) &0.403(0.05)  & $\textcolor{blue}{0.422(0.01)}$  \\ 

    \hline
            
    \multirow{3}*{PB} &  PLC     & 0.622(0.04) & 0.531(0.09)  &$\textcolor{blue}{0.638(0.04)}$  & 0.627(0.07) \\ 
                        &  SRC     & 0.583(0.05) & 0.575(0.07) &0.625(0.05) & $\textcolor{blue}{0.627(0.07)}$ \\ 
                         &  KLC     & 0.420(0.04) & 0.408(0.05) &0.443(0.04)  & $\textcolor{blue}{0.445(0.06)}$  \\ 
    \hline

    \multirow{3}*{AA} &  PLC     & 0.771(0.01) & 0.793(0.01) &0.785(0.02)  & $\textcolor{blue}{0.801(0.01)}$ \\ 
                    &  SRC     & 0.796(0.03) & 0.793(0.00) &0.799(0.02)  & $\textcolor{blue}{0.813(0.01)}$  \\ 
                    &  KLC     & 0.610(0.03) & 0.609(0.01) &0.614(0.02)  & $\textcolor{blue}{0.628(0.01)}$  \\ 
    \toprule[1pt]
    \end{tabular}
\end{table}
Table~\ref{abl_table} shows the ablation experiments on SRC loss. Res-$\mathrm{SRC_{ascent}}$ indicates that only the ascent part in SRC loss was used, as shown in Eq.~\ref{spearman_loss_part2_pos}, and Res-$\mathrm{SRC_{descent}}$ is the opposite. Experiments show that constraining both the ascent and descent parts simultaneously can achieve the best results.  Fig.~\ref{distribute_fig} illustrates the consistency and correlation distribution map of three datasets in two regression tasks. Our proposed methods can reduce outliers very well and achieve excellent performance on correlation and consistency in various tasks.\par 
\par
    
\section{Conclusion}
In this paper, we propose two correlation-based loss functions for medical image regression tasks. By constraining the outliers and explicitly regularizing the key distribution factors of normal samples, our proposed PLC loss exhibits powerful capabilities on regression tasks. Moreover, we propose a Coarse-to-Fine optimization strategy to ease the rank learning, which can further improve regression performance. Experimental results show that the simple network equipped with our proposed loss functions can achieve excellent performance on various medical image regression tasks.\par
\subsubsection{Acknowledge.}
This work was supported by the grant from National Natural Science Foundation of China (Nos. 62171290, 62101343), Shenzhen-Hong Kong Joint Research Program (No. SGDX20201103095613036), and Shenzhen Science and Technology Innovations Committee (No. 20200812143441001).

\bibliographystyle{splncs04}
\bibliography{paper1280.bib}

\end{document}